\documentclass{article}
\usepackage[colorlinks]{hyperref}
\usepackage{spconf,amsmath,graphicx}
\usepackage[capitalize]{cleveref}
\crefname{section}{Sec.}{Secs.}
\crefname{table}{Table}{Tables}
\crefname{figure}{Fig.}{Figs.}
\usepackage{adjustbox}
\usepackage{xcolor}
\usepackage{multirow}
\usepackage{amsmath}

\DeclareMathOperator*{\argmax}{arg\,max}

\title{Open-Vocabulary Panoptic Segmentation Using BERT Pre-Training of Vision-Language Multiway Transformer Model}
\name{Yi-Chia Chen\textsuperscript{*}\thanks{*Equal Contribution.}, Wei-Hua Li\textsuperscript{*}, Chu-Song Chen}
\address{National Taiwan University}

\begin{document}

\maketitle

\begin{abstract}
Open-vocabulary panoptic segmentation remains a challenging problem. One of the biggest difficulties lies in training models to generalize to an unlimited number of classes using limited categorized training data. Recent popular methods involve large-scale vision-language pre-trained foundation models, such as CLIP. In this paper, we propose OMTSeg for open-vocabulary segmentation using another large-scale vision-language pre-trained model called BEiT-3 and leveraging the cross-modal attention between visual and linguistic features in BEiT-3 to achieve better performance. Experiments result demonstrates that OMTSeg performs favorably against state-of-the-art models. Code is available at \href{https://github.com/AI-Application-and-Integration-Lab/OMTSeg}{https://github.com/AI-Application-and-Integration-Lab/OMTSeg}
\end{abstract}

\begin{keywords}
Open vocabulary, panoptic segmentation.
\end{keywords}

\section{Introduction}
\label{sec:intro}
Image segmentation is a 
task that involves dividing the 
input image into regions corresponding to different objects, which can be used in various scenarios~\cite{liu2023optimizing, gerstenberger2023differentiable}. 
Recently, open-vocabulary image segmentation has attracted more attention and remains a challenging task. This task focuses on segmenting an image into instance-based or semantically consistent regions according to arbitrary text descriptions.
The segmentation category labels have not necessarily been seen in the training phase. There are different approaches tackling this problem, but one of the most promising solutions is to leverage pre-trained large-scale vision-language models. CLIP~\cite{radford2021learning} connects text and images using contrastive learning by aligning the semantic meanings of images and texts in a shared vector space, which was the mostly adopted one that serves as a fundamental knowledge base.
It is then upgraded from whole-image-level descriptions to region-level panoptic segmentation for open-vocabulary image segmentation.

A main advantage of the CLIP model over traditional computer vision models is that it can perform zero-shot learning, which means it can classify images into categories that it has never seen before, by simply using natural language descriptions of the categories.
Hence, it can handle open-ended tasks, which means it can answer any natural language query about an image, not just predefined ones.
However, CLIP uses a contrastive learning framework that consists of two components: an image encoder and a text encoder, which are independent of each other and are only aligned on the final output space of the feature descriptions. 
Therefore, CLIP does not fully exploit the cross-reference clues between the vision and language modalities in its middle-layer latent representations, restricting its open-vocabulary generalization performance on vision-language 
or pure vision tasks.


Specifically, to achieve open vocabulary image segmentation, a common method is to use a foundation vision-language model that can connect the visual concepts and natural language for an entire image input as a knowledge base. Additional structure added to the knowledge-base model then forms a downstream model, which is used to learn from the data with segmentation labels for only a few classes of objects. Inspired by these limited examples, the downstream model can then learn to segment and label any object, regardless of whether the category has been seen before.

However, simply applying the output-layer embedding of the text and visual encoders of CLIP would be insufficient. To enforce the open-vocabulary capability by learning from limited segmentation examples, layer-wise latent representations have been adopted in several recent studies, e.g., MaskCLIP~\cite{ding2023maskclip} and SAN~\cite{xu2023side}. A main issue is that the CLIP model lacks of considering the interaction between the latent representations of the image encoder and text encoder, so the intermediate representation it learns becomes less representative, restricting the performance of employing hierarchical latent representations for open-vocabulary segmentation.

Therefore, we believe that cross-referred latent representations between the vision and language modalities are crucial for generalizing the image description capability from the whole-image level to the local-regional level. To leverage better the cross-modality information, we introduce an open-vocabulary image segmentation method leveraging the multi-modal foundation model based on BERT pre-training of multiway transformers~\cite{beit3, vlmo}. 

Our approach, Open-vocabulary Multiway Transformer Segmentation (OMTseg), is simple but effective for open vocabulary segmentation.
OMTseg is established based on BEiT-3~\cite{beit3}, which is a multiway transformer model that has been learned with comprehensive corss-modality information between the layerwise representations. Our approach can therefore takes advantage of more complete cross-reference latent information between vision and language models, benefiting to extract regional clues from the whole image.

BEiT-3 has been used for traditional image segmentation where the training and testing class labels are the same, and achieves remarkable performance. However, only the vision module of BEiT-3 is used as a pre-trained model for the downstream image segmentation tasks in the above study. To achieve open-vocabulary image segmentation, a crucial issue is to employ also the language module that can identify the highly probable potential classes in an image. In our approach, we feed the respective language-related embeddings to the vision module and seek for a coincide segmentation result. This is achieved through the collaborative cooperation of both vision and language modules, instead of a single-modality model in the traditional image segmentation.

In this work, we introduce OMTSeg, a novel approach for open-vocabulary image segmentation. Our method is rooted in the integration of the BEiT-3 model, a multiway transformer adept in processing large-scale vision-language data. The distinctive features of OMTSeg include:

\noindent \textbf{Cross-Modal Attention}: Utilizing cross-modal attention within BEiT-3, OMTSeg effectively combines visual and linguistic inputs, providing segmentations of complex scenes.

\noindent \textbf{Layer-wise Latent Representations}: OMTSeg leverages latent representations of both vision and language encoders, 
ensuring 
extensive integration of cross-modal information.

\noindent \textbf{Visual Adapter and Language Prompting}: OMTSeg incorporates a visual adapter to enhance the capabilities of the BEiT-3 backbone and leverages a language prompting mechanism that refines text embeddings to align with visual context and enhances segmentation.

OMTSeg processes images and text through BEiT-3, integrates features via cross-modal attention, and performs segmentation and classification using a multiway segmentation head. This approach allows for accurate segmentation of both seen and unseen categories, marking a significant advancement in open-vocabulary image segmentation.

\section{Related Works}
Semantic Segmentation and Instance Segmentation are two fundamental tasks in image-based content analysis. Panoptic Segmentation performs both semantic and instance segmentation in an image to achieve a comprehensive segmentation result, which is a field that bridges semantic-level and instance-level visual clues to conduct image segmentation. Recent studies in sementatic segmentation include SegFormer~\cite{xie2021segformer} and Segmenter~\cite{strudel2021segmenter}, which employ structured Transformer encoders or Vision Transformer (ViT) to conduct efficient approaches for per-pixel classification. Outstanding approaches in panoptic segmentation include MaskFormer~\cite{cheng2021maskformer}, which introduces a unified model to address both tasks using a mask classification approach. This model outperforms traditional per-pixel classification methods, especially when dealing with a large number of classes. Moreover, Mask2Former~\cite{cheng2021mask2former} presents a Masked-attention Mask Transformer that can handle various image segmentation tasks, reducing research effort and setting new benchmarks for panoptic segmentation. 

Our approach, on ther other hand, handles the open-vocabulary segmentation where the testing labels can be arbitrary.
Since our approach leverages vision-language foundation models and performs panoptic segmentation for open-vocabulary labels, we adopt Mask2Former~\cite{cheng2021mask2former} as our segmentation header that undertakes the foundation model outputs and extend it to an open-vocabulary panoptic segmention model. In the following, we first give a review of recent vision-language foundation models. Then, we survey the works of open-vocabulary image segmentation. Finally, we briefly introduce the multiway transformers.
\vspace{-10pt}
\subsection{Vision-Language Foundation Models}
The recent development of large-scale vision-language models involves the integration of visual and linguistic features, trained on vast amounts of data, resulting in powerful capabilities.
Early works like 
VinVL~\cite{zhang2021vinvl} 
lays the foundation. It was later expanded by models like CLIP~\cite{radford2021learning},
which enhances scalability and widening the applicability across numerous tasks. 
BLIP and BLIP-2~\cite{li2023blip} have bootstrapped language-image pre-training, showing improved capabilities in image captioning. 
BEiT-3~\cite{beit3} further contributes by integrating vision self-supervised learning, significantly boosting performance in visual tasks. 
These models represent a shift towards building robust and scalable solutions for a wide range of vision-language tasks.

\subsection{Open Vocabulary Image Segmentation}
Open vocabulary segmentation has made significant contributions with the introduction of SimSeg [12], which leverages CLIP for semantic segmentation.
Following models like OVSeg~\cite{liang2023open} and MaskCLIP~\cite{ding2023maskclip} either fine-tunes CLIP or introduces novel architectures to better manage masked images. 
ODISE~\cite{xu2023odise} brings together pre-trained text-image diffusion and discriminative models for open-vocabulary panoptic segmentation. SAN~\cite{xu2023side} introduces a Side Adapter Network for region recognition and FC-CLIP~\cite{yu2023fcclip} simplifies the segmentation pipeline into a single-stage framework, enhancing the balance between accuracy and cost. 
These efforts emphasize the importance of using pre-trained vision-language models and innovative architectures to improve open-vocabulary segmentation. 

\begin{figure*}[!ht]
    \centering
    \includegraphics[width=1.75\columnwidth]{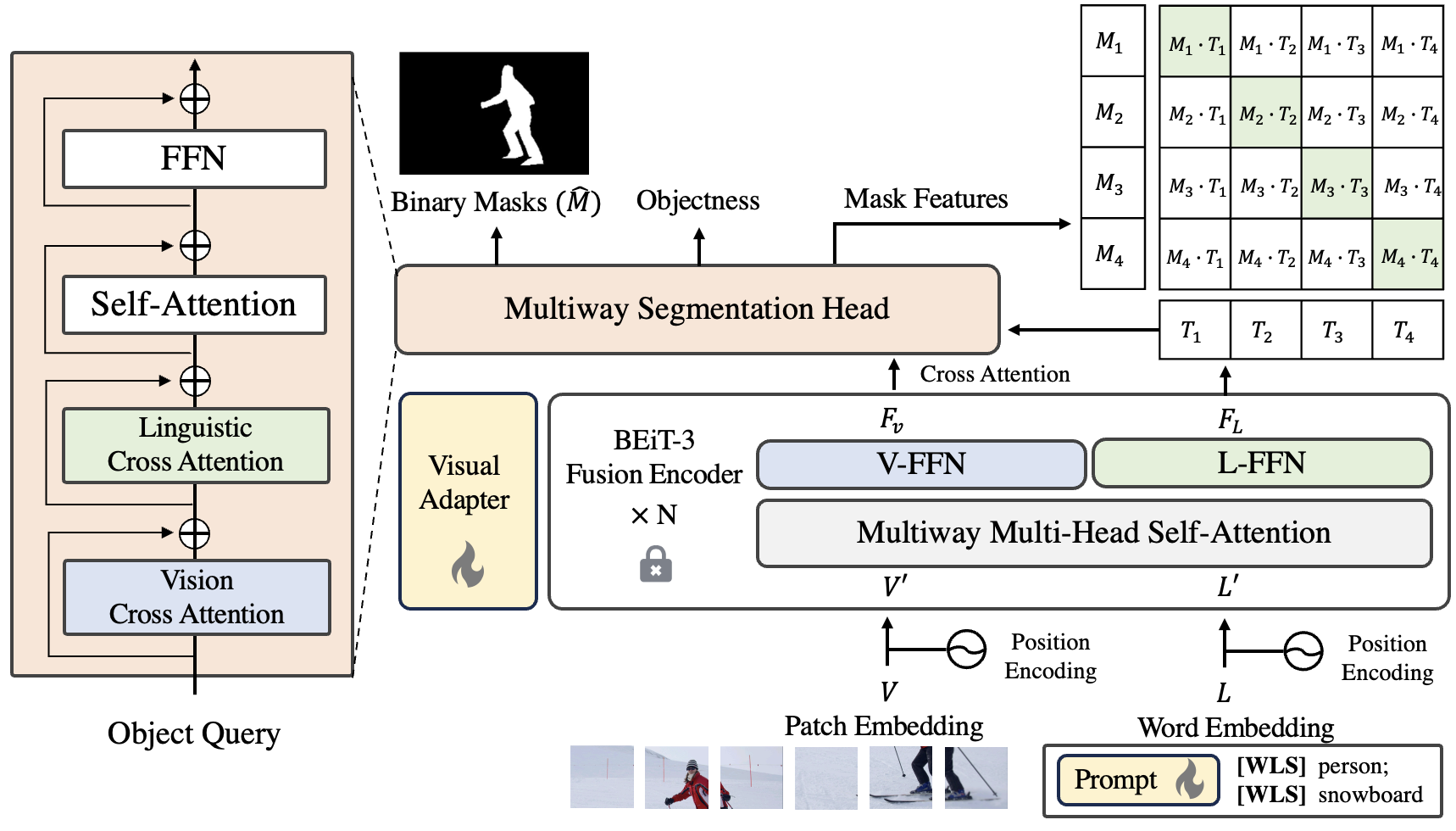}
    \caption{Overview of OMTSeg, which contains BEiT-3 fusion encoder and a multiway segmentation head.}
    \label{fig:overview_img}
    \vspace{-10pt}
\end{figure*}

\subsection{Multiway Transformers}
Recently, vision-language and multimodal learning have 
become popular and 
yielded various models.
A noteworthy direction is the development of multiway transformers. 
VLMo~\cite{vlmo} introduces the Mixture-of-Modality-Experts Transformer enabling both dual-encoder and fusion-encoder capabilities within a single architecture.
The model leverages stagewise pre-training to efficiently utilize both image-only and text-only data. 
BEiT-3~\cite{beit3}, on the other hand, moves a step further by introducing a general-purpose multimodal foundation model. 
It employs multiway transformers that can adapt to various modalities, including images (``Imglish") and texts (English), under a single pre-training task.

\vspace{-8pt}
\section{Proposed Method}
\label{sec:method}
The task of Open Vocabulary Panoptic Segmentation (OVPS) involves segmenting an input image $I$ into a set of regions, each associated with a category label and an instance identifier. The challenge lies in handling both seen and unseen categories. Let $C_{\text{seen}}$ denote the set of categories seen during training and $C_{\text{unseen}}$ the set of categories not encountered during training. The goal is to accurately segment and classify each pixel in $I$ into one of the categories in $C=C_{\text{seen}} \cup C_{\text{unseen}}$ and assign instance identifiers for object instances. 
Formally, for each pixel $p$ in the image $I$, an OVPS model aims to output a tuple $(c_p, i_p)$, where $c_p \in C$ is the category label and $i_p$ is the instance identifier. The instance identifier $i_p$ is unique for each object instance and is consistent for `stuff' categories i.e., uncountable and amorphous regions like grass, sky, etc.

\subsection{OMTSeg}
In this section, we provide an in-depth description of our OMTSeg model. 
Fig.~\ref{fig:overview_img} depicts an overview of OMTSeg. 
The input comprises an image and a text string. 
The image is in RGB format $I \in R^{H \times W \times 3}$ with $H$ and $W$ the height and width of the image, respectively. 
The text input consists of a sequence of category names, each associated with a special token \textbf{[WLS]} to split the category text from the text input and indicate the features corresponding to that category.
E.g., if the target categories include `person', `snow', and `snowboard', the text input is formatted as 
\begin{equation}
\textbf{[WLS]} \, \text{person}\; ; \; \textbf{[WLS]} \, \text{snow}\; ; \; \textbf{[WLS]} \, \text{snowboard}.
\end{equation}
Next, we apply tokenization and word embedding to generate $T_\text{input}\in R^{N_\text{w}}$, where $N_\text{w}$ represents the number of the total word in text input. 
This structure of text input allows the model to effectively link the visual categories with the linguistic descriptions provided in the text. 

\subsubsection{BEiT-3 Backbone}
BEiT-3 is employed as a core feature extraction backbone in OMTSeg, which integrates visual and linguistic inputs. First, The image input $I$ is divided into patches, each linearly transformed into an embedding vector to form a visual feature $V \in R^{N_p \times d}$, where $N_p$ represents the number of patches and $d$ represents the dimension of each token. The text input $T_\text{input}$ is similarly encoded into a sequence of embeddings $L \in R^{N_\text{w} \times d}$, representing the linguistic features. To obtain spatial and sequential context, positional encodings are added to both visual ($V$) and linguistic ($L$) embeddings, resulting in enhanced feature sets ($V'$) and ($L'$), which can be represented as $[{V'}, {L'}]= [{V}, {L}] + \text{Positional Encoding}$.

Our backbone consists of a series of multiway transformer layers. 
We apply $N$ layers of fusion encoder proposed in BEiT-3, where each layer processes the concatenated visual and linguistic embeddings through the operations of Multiway Multihead Self-Attention (MultiwayMHSA) and Layer Normalization (LN), followed by an independent Feed-Forward Network for visual (V-FFN) and linguistic feature (L-FFN). The multiway transformer layer operates as follows:
\begin{equation}
\resizebox{0.42\textwidth}{!}{$
\begin{aligned}
    &[{V'}, {L'}] = \text{LN}(\text{MultiwayMHSA}([{V'}, {L'}]) + [{V'}, {L'}]) \\
    & F_V, F_L= \text{LN}(\text{V-FFN}(V') + V'), \text{LN}(\text{L-FFN}(L') + L')\\
\end{aligned}
$}
\end{equation}
This structure, repeating across multiple layers, facilitates the 
integration of vision and language features, producing a rich, joint representation for segmentation tasks.

\begin{figure}[!t]
    \centering
    \includegraphics[width=1\columnwidth]{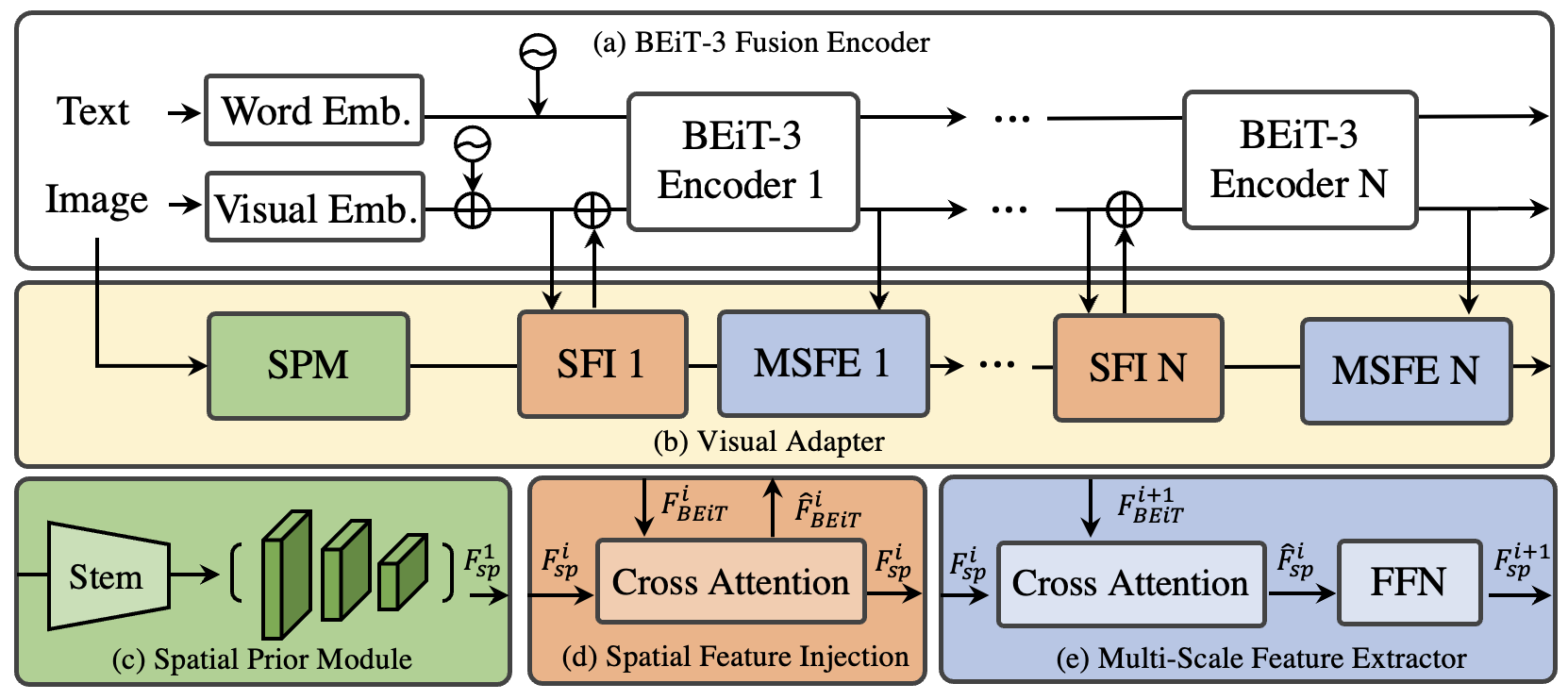}
    \caption{Overall architecture of Visual Adapter. (a) BEiT-3 Encoder for feature extraction and integrates visual and linguistic feature.(b) Visual adapter is constructed by stacking SPM, SFI, and MSFE. (c) SPM (d) SFI (e) MSFE.}
    \vspace{-5pt}
    \label{fig:visual_adapter}
\end{figure}

\subsubsection{Visual Adapter for BEiT-3}
In the OMTSeg model, the Vision Adapter is specifically designed to enhance the capabilities of the frozen BEiT-3 backbone, making it more adept at handling dense predictions. Fig.~\ref{fig:visual_adapter} illustrates the overall architecture of the visual adapter. Following~\cite{chen2022vision}, the Vision Adapter consists of three main components: Spatial Prior Module
 (SPM), Spatial Feature Injector (SFI), and Multi-Scale Feature Extractor (MSFE), each tailored to complement the BEiT-3 architecture.
\noindent\textbf{SPM} operates in parallel with BEiT-3's patch embedding layer, capturing local spatial contexts of images. It utilizes a convolutional stem similar to that in ResNet, comprising a series of convolutions and a max-pooling layer. This is followed by stride-2 3x3 convolutions and 1x1 convolutions, which project the feature maps to $d$ dimensions. 
SPM is designed to capture spatial features at various resolutions, including 1/8, 1/16, and 1/32 of the original image size.
\noindent\textbf{SFI} bridges the spatial features from SPM with the features extracted by BEiT-3. For each block of BEiT-3, the spatial features ${F}_{\text{sp}}^{i}$ act as keys and values in a cross-attention mechanism 
and the 
features ${F}_{\text{BEiT}}^{i}$ 
serve as queries. This module effectively injects spatial context into the BEiT-3 features as
\begin{equation}
    {\hat{F}}_{\text{BEiT}}^{i} = F_{\text{BEiT}}^{i} + \text{CrossAttention}({F}_{\text{BEiT}}^{i}, {F}_{\text{sp}}^{i}).
\end{equation}

\noindent\textbf{MSFE} further processes these features to produce multi-scale representations. 
It combines cross-attention and an FFN, enhancing the feature extraction capabilities for varied scales:

\begin{equation}
    {F}_{\text{sp}}^{i+1} = {\hat{F}}_{\text{sp}}^{i} + \text{FFN}(\text{LN}(\hat{{F}}_{\text{sp}}^{i}))
\end{equation}
\begin{equation}
    \hat{{F}}_{\text{sp}}^{i} = {F}_{\text{sp}}^{i} + \text{Attention}(\text{LN}({F}_{\text{sp}}^{i}), \text{LN}({F}_{\text{BEiT}}^{i+1}))
\end{equation}

The incorporation of Vision Adapter into BEiT-3 allows for effective adaptation of the transformer architecture for dense prediction tasks, significantly enhancing its 
multi-scale and location-aware feature representation capabilities.

\subsubsection{Language Prompting}
\label{subsec:language_prompt}

In OMTSeg, the Language Prompt plays a crucial role. It involves tuning the embeddings of special tokens, namely the \textbf{[WLS]} tokens, to better align the linguistic features with the visual context. Prompt tuning in OMTSeg focuses on the \textbf{[WLS]} tokens, which are designed to encapsulate category-specific information. These tokens undergo an adaptation process whose embeddings are fine-tuned to capture the nuances of each category. The tuning process is as follows:
\begin{equation}
    \mathbf{Embed}_{\text{wls}} \leftarrow \mathbf{Embed}_{\text{wls}} + \Delta\mathbf{Embed}
\end{equation}
Here, \(\mathbf{Embed}_{\text{wls}}\) represents the initial embedding of the \textbf{[WLS]} token derived from the embedding of the \textbf{[CLS]} token in the transformer model. \(\Delta\mathbf{Embed}\) is the trainable adjustment applied to the embedding, allowing each \textbf{[WLS]} token to uniquely represent its corresponding category.
The fine-tuned \textbf{[WLS]} token embeddings are then integrated with the visual features from the BEiT-3 backbone. This integration ensures that the model not only captures the visual characteristics of each category but also aligns them with the corresponding textual descriptions. 
The Language Prompt mechanism in OMTSeg serves as a bridge between language and vision, enabling the model to leverage textual cues for enhanced segmentation performance.

\subsubsection{Multiway Segmentation Head}
\label{subsec:segmentation_head}

Multiway Segmentation Head in OMTSeg is a critical component for producing segmentation masks, integrating the processed visual and linguistic features. 

The architecture of the multiway segmentation head is inspired by Mask2Former~\cite{cheng2021mask2former}, featuring multiple transformer decoders equipped with visual and linguistic cross attention to enhanced integration of visual and linguistic features.
The architecture includes Masked Multi-Head Cross Attention (MaskedMHCA), Multi-Head Cross Attention (MHCA), Multi-Head Self-Attention (MHSA) and 
FFN.
The Transformer decoder layer in the Multiway Segmentation Head can be represented as follows:
\begin{align}
     {X} &= \text{MHCA}(\text{MaskedMHCA}({X}, {F_V}), {F_L}),\\
     {O} &= \text{FFN}({\text{MHSA}({X}))},
\end{align}
where $X, F_V, F_L, O$ represent object query features, visual features, textual features and the output respectively. 
The output of the segmentation head contains (1) a set of binary mask ($\hat{M}$), each corresponding to an identified region in the image, (2) mask feature ($F_M$), which is the embedding feature corresponding to each mask, (3) objectness value, to indicate whether the corresponding mask has an object.

Regarding the loss function of OMTSeg, we utilize the binary cross-entropy loss for our mask loss and objective loss. Besides, we also apply contrastive loss proposed in~\cite{radford2021learning} to align mask feature and linguistic feature.

\subsection{Open-Vocabulary Classification}
Unlike traditional close-set segmentation approaches, our method leverages a feature cosine similarity mechanism for classification tasks. This technique allows for a more flexible and descriptive segmentation process, accommodating a wide array of classes without the need for predefined 
categories. Given an mask feature vector $F_M$ and a corresponding linguistic feature $F_L$, the classification is performed via the cosine similarity between these two vectors. The class label is then assigned by $\textit{class} = \underset{c}{\argmax} \cos(F_M, F_L)$, where $\cos$ denotes the cosine similarity function, and $c$ ranges over all possible classes. This formulation enables the model to match image segments with their most relevant linguistic descriptors and subsequently implements open-vocabulary classification. 

\begin{table*}[!ht]
\centering
\caption{Open-vocabulary semantic segmentation performance on ADE20K-847 (A-847), Pascal Context-459 (PC-459), ADE20K-150 (A-150), Pascal Context-59 (PC-59), Pascal VOC (PAS-21, PAS-20). The best score is in bold, and the second best is underlined. ``-" means the score is not released/available. \dag \ means the model is trained with caption data.}
\vspace{.5em}
\label{tab:main_results}
\begin{adjustbox}{width=2\columnwidth,center}
\begin{tabular}{|c|c|c|c|c|c|c|c|}
\hline

\multirow{2}{*}{Method} & \multirow{2}{*}{Training Data} & \multicolumn{6}{c|}{mIoU}\\ \cline{3-8}
                    &    & A-847& PC-459& A-150& PC-59& PAS-21& PAS-20       \\ \hline
ZS3Net (NeurIPS'19)                       & Pascal VOC                   & -             & -             & -             & 19.4          & 38.3          & -             \\ \hline
LSeg (ICLR'22)                   & Pascal VOC                   & -             & -             & -             & -             & 47.4          & -             \\ \hline
GroupViT (CVPR'22)                & GCC+YFCC                     & 4.3           & 4.9           & 10.6          & 25.9          & 50.7          & 52.3          \\ \hline
SimSeg (CVPR'23)                  & COCO Stuff                   & -             & -             & 15.3          & -             & 74.5          & -             \\ \hline
ZegFormer (CVPR'22)                  & COCO Stuff                   & -             & -             & 16.4          & -             & 73.3          & -             \\ \hline
LSeg+ (ICLR'22)             & COCO Stuff                   & 3.8           & 7.8           & 18.0          & 46.5          & -             & -             \\ \hline
OVSeg (CVPR' 23)          & COCO Stuff                   & 9.0           & 12.4          & 29.6          & 55.7          & -             & \underline{94.5}          \\ \hline
SAN (CVPR'23)& COCO Stuff                   & {\underline {13.7}}    & \textbf{17.1} & {\underline{33.3}}    & {\underline{60.2}}    & -             & \textbf{95.5} \\ \hline
OpenSeg (ECCV'22)           & COCO Panoptic + COCO Caption & 6.3           & 9.0           & 21.1          & 42.1          & -             & -             \\ \hline
ODISE\dag(CVPR'23)     & COCO Panoptic + COCO Caption & 11.0          & 13.8          & 28.7          & 55.3          & 82.7          & -             \\ \hline
ODISE (CVPR'23)          & COCO Panoptic                & 11.1          & \underline{14.5}          & 29.9          & 57.3          & {\underline{84.6}}    & -             \\ \hline
MaskCLIP (ICML'23)           & COCO Panoptic                & 8.2           & 10.0          & 23.7          & 45.9          & -             & -             \\ \hline
OMTSeg (Ours)                 & COCO Panoptic                & \textbf{13.9} & \textbf{17.1} & \textbf{34.8} & \textbf{61.0} & \textbf{85.2} & \textbf{95.5} \\ \hline
\end{tabular}
\end{adjustbox}
\vspace{-15pt}
\end{table*}

\begin{table}[!ht]
\centering
\caption{
Panoptic segmentation performance. \dag \ means the model is trained with caption data.}
\vspace{.5em}
\label{tab:ade20k_results}
\begin{adjustbox}{width=1\columnwidth,center}

\begin{tabular}{cccc}
\hline
\multirow{2}{*}{Method} & \multirow{2}{*}{Params (M)} & ADE20K           & COCO             \\
                        &                             & PQ/AP/mIoU       & PQ/AP/mIoU       \\ \hline
MaskCLIP              & 367                         & 15.1/ 6.0/ 23.7  & -/ -/ -          \\
FreeSeg             & -                           & 16.3/ 6.5/ 24.6  & -/ -/ -          \\
ODISE                 & 1522                        & 22.6/ \underline{14.4}/ \underline{29.9} & \textbf{55.4}/ \textbf{46.0}/ \textbf{65.2} \\
ODISE\dag               & 1522                        & \underline{23.4}/ 13.9/ 28.7 & 45.6/ 38.4/ 52.4 \\
OMTSeg(Ours)            & 720                         & \textbf{27.5}/ \textbf{17.4}/ \textbf{34.8} & \underline{54.9}/ \underline{45.0}/ \underline{64.0} \\ \hline
\end{tabular}
\end{adjustbox}
\vspace{-5pt}
\end{table}

\section{Experiments}

In this section , we present the experiments to evaluate 
the proposed OMTSeg for open-vocabulary 
segmentation. 

\subsection{Datasets and Evaluation Criteria}
In our experiments, we employ 
six benchmark datasets to thoroughly validate our approach. These include the datasets of COCO Panoptic, ADE20K-150, ADE20K-847, Pascal Context-59, Pascal Context-459, and Pascal VOC. 

\noindent\textbf{COCO Panoptic} is an extension of the original COCO dataset. It serves as our primary training ground owing to its extensive class diversity and size.

\noindent\textbf{ADE20K-150} is a subset of 
ADE20K~\cite{zhou2019semantic}, 
which includes 150 classes. It provides a balanced mix of object and stuff categories, thereby enabling a comprehensive evaluation.

\noindent\textbf{ADE20K-847} is the full version of the ADE20K dataset, consisting of 847 categories. It offers a more challenging evaluation scenario. 

\noindent\textbf{Pascal Context-59} 
is derived from the original Pascal VOC, 
featuring 59 context-aware classes. It allows for an evaluation setting where context plays a significant role.

\noindent\textbf{Pascal Context-459} is an expanded version of Pascal Context-59. This dataset contains 459 classes and provides a more exhaustive setting for evaluation.

\noindent\textbf{Pascal VOC} 
is one of the classic datasets in object segmentation. {\color{black}``PAS-20" in Tabel~\ref{tab:main_results} denotes the Pascal VOC dataset with 20 object categories, while ``PAS-21" indicates a version with 21 categories, including a further background class.}

Regarding the Evaluation Criteria, the metrics we employ are Average Precision, mean Intersection over Union, and Panoptic Quality. \textbf{Average Precision (AP)} is a commonly used metric to evaluate the performance of a segmentation model. AP is calculated as the area under the precision-recall curve. It measures the precision of the model at different recall levels. \textbf{Mean Intersection over Union (mIoU)} reflects the overall quality across all classes. A higher mIoU indicates better performance in capturing the agreement between the predicted and ground truth segmentation masks. \textbf{Panoptic Quality (PQ)} considers two main aspects: semantic segmentation quality and instance segmentation quality. It provides a comprehensive measure of the overall performance of a panoptic segmentation model.

\subsection{Implementation Details}
We conduct our experiments using 
BEiT-3-large models as the backbone architectures. For training, we utilize the Lion optimizer 
with a weight decay of \(0.15\). The input images are cropped to a fixed size of \(640 \times 640\). The learning rate is set to \(3 \times 10^{-5}\) and the model is trained for 90,000 iterations with a batch size of 16. The first 600 steps are used for warm-up, followed by a linear decay of the learning rate for the remaining iterations.

\subsection{Experiment Setting and Results}

\textbf{Open-Vocabulary Semantic Segmentation:} To evaluate the performance of open-vocabulary segmentation, we train our OMTSeg model using the COCO Panoptic training set and perform evaluations 
on the remaining benchmark datasets 
including ADE20K, COCO, Pascal Context, and Pascal VOC.
Following~\cite{xu2023odise}, we utilize mIoU to evaluate the performance 
on semantic segmentation. 
We compare our OMTSeg with 
numerous methods including 
ZS3Net~\cite{bucher2019zero}, LSeg~\cite{Li2022LanguagedrivenSS}, GroupViT~\cite{xu2022groupvit}, SimSeg~\cite{yi2023simple}, ZegFormer~\cite{ding2022decoupling}, OVSeg~\cite{liang2023open}, SAN~\cite{xu2023side}, OpenSeg~\cite{ghiasi2022scaling}, ODISE~\cite{xu2023odise} and MaskCLIP~\cite{ding2023maskclip}.
As can be seen in Table~\ref{tab:main_results}, OMTSeg achieves 
more favorable performance against previous methods, which reveals that our model of integrating vision adaptor and text prompt tuning into the cross-attention mechanism can effectively boost the open-vocabulary segmentation performance.

\noindent\textbf{Panoptic Segmentation:} To evaluate the performance on panoptic segmentation, we train OMTSeg on the COCO Panoptic training set and evaluate it on ADE20K and COCO datasets, where these two datasets are commonly employed for the performance evaluation of panoptic segmentation.
We use PQ, AP and mIoU to measure the performance for panoptic, instance, and semantic segmentations, respectively. 
As shown in Table~\ref{tab:ade20k_results}, on the ADE20K dataset, OMTSeg achieves a PQ score of 27.5, surpassing the previous state-of-the-art models. 
This reveals OMTSeg's capability to 
recognize and segment complex scenes of unseen classes. 
On the COCO dataset, OMTSeg shows competitive performance with a PQ score of 54.9, evidencing its robustness across diverse situations (where COCO is a close-set testing). 
Although our 
method performs slightly 
worse than ODISE~\cite{xu2023odise}, our model's 
size (number of parameters) is considerably smaller. 
\vspace{-15pt}
\subsection{Ablation Studies}
We 
conduct ablation study on the component of OMTSeg, include Cross Attention, Visual Adapter and Text Prompt Tuning 
on the ADE20K dataset. The results are shown in Table~\ref{tab:ablation}. 

\noindent\textbf{Cross Attention:} Removal of the cross-attention mechanism resultes in a significant drop in performance (PQ score from 27.5 to 18.4). 
This highlights its crucial role in the effective integration of vision and language modalities. 

\noindent\textbf{Visual Adapter:} Excluding the visual adapter leads to a marked performance decrease, which indicates the adapter's importance in augmenting the base transformer model for accurate segmentation.

\noindent\textbf{Text Prompt Tuning:} Without text prompt tuning, the model's PQ score is decreased to 25.9, emphasizing the value of fine-tuning text prompts for better alignment between linguistic features and visual context.

\begin{table}[!t]
    \centering
    \caption{Ablation Studies on Cross Attention, Visual Adapter and Text Prompt Tuning.}
    \vspace{.5em}
    \begin{adjustbox}{width=0.75\columnwidth,center}
    \begin{tabular}{llll}
\hline
                           & \multicolumn{3}{c}{ADE20K} \\
                           & PQ      & AP      & mIoU   \\ \hline
\multicolumn{1}{l}{OMTSeg} & \textbf{27.5} & \textbf{17.4}    & \textbf{34.8}   \\ \hline
w/o text cross-attn.       & 18.4    & 9.1     & 21.1   \\
w/o visual adapter         & 8.8     & 3.0     & 12.5   \\
w/o text prompt tuning     & \underline{25.9}    & \underline{16.6}    & \underline{30.4}   \\ \hline
\end{tabular}
    \end{adjustbox}
    \label{tab:ablation}
\end{table}

\vspace{-5pt}
\section{Conclusion}
We introduced the Open Vocabulary Multiway Transformer Segmentation (OMTSeg), which leverages BEiT-3 with several architectural advancements. 
Our method is simple but effective and achieves state-of-the-art performance across various datasets. 
The integration of cross-modal attention, visual adapters, and language prompt tuning has proven to be 
effective, as evidenced by our comprehensive experimental evaluations and ablation studies. 
Future work will explore the extension of OMTSeg to other vision-language tasks. 
The results of OMTSeg pave the way for more advanced and efficient open-vocabulary segmentation models. 

\noindent\textbf{Acknowledgement:}
This work was supported in part by the National Science and Technology Council, Taiwan under Grants NSTC 112-2634-F-002-005 \& 112-2221-E-002-182-MY3, NTU under grants 113L900902, and in part by an unrestricted gift from Google. We thank to National Center for High-performance Computing (NCHC) of National Applied Research Laboratories (NARLabs) in Taiwan for providing computational and storage resources.

\bibliographystyle{IEEEbib}
\bibliography{strings,refs}

\end{document}